# Prompt fidelity of ChatGPT4o / Dall-E3 text-to-image visualisations

*Dirk HR Spennemann*

School of Agricultural, Environmental and Veterinary Sciences, Charles Sturt University; Albury NSW 2640, Australia

Libraries Research Group; Charles Sturt University; Wagga Wagga NSW 2678, Australia

Correspondence: dspennemann@csu.edu.au

**Abstract**

This study examines the prompt fidelity of ChatGPT4o / DALL-E3 text-to-image visualisations by analysing whether attributes explicitly specified in autogenously generated prompts are correctly rendered in the resulting images. Using two public-domain datasets comprising 200 visualisations of women working in the cultural and creative industries and 230 visualisations of museum curators, the study assessed accuracy across personal attributes (age, hair), appearance (attire, glasses), and paraphernalia (name tags, clipboards). While correctly rendered in most cases, DALL-E3 deviated from prompt specifications in 15.6% of all attributes (n=710). Errors were lowest for paraphernalia, moderate for personal appearance, and highest for depictions of the person themselves, particularly age. These findings demonstrate measurable prompt-to-image fidelity gaps with implications for bias detection and model evaluation.

**Keywords**

artificial intelligence; algorithmic bias; large language models; text-to-image; prompt conversion; prompt adherence; prompt compliance; visualization accuracy

## 1. Introduction

A considerable body of work has examined the biases inherent in text-to-image generative Artificial Intelligence (AI) models ever since ChatGPT came onto the scene in November 2022. A wide range of biases, primarily in the form of stereotypes, have been identified, mainly in misrepresenting gender (favouring males), ethnicity (Caucasian), age (young) and beauty [1-6]. Images of pregnant women or people with disabilities or degenerative diseases, on the other hand, are largely absent in single-shot Ai-generated prompts [3] or are misrepresented [7, 8]. Similarly, biases have been observed in image backgrounds of generative AI portraits [9, 10].

Since its public release of GPT3.5 by OpenAI on 30 November 2022 [11, 12], ChatGPT has undergone a range of improvements (setting aside various sub-models), with ChatGPT4 released in March 2023 [13] and ChatGPT4o in May 2024 [14]. The latest version, ChatGPT5, was released in August 2025 [15]. While ChatGPT3.5 was text-to-text only, with any text-to-image processing carried out using the standalone application DALL-E2 [16] and later DALL-E3 [17], ChatGPT4 accepted user-defined image generation prompts that were interpreted, with ChatGPT4 autogenously generating prompts for parsing into DALL-E3. In March 2025 OpenAI replaced the diffusion model DALL-E3 with a ChatGPT native image generation model as the text-to-image engine that is interfaced with ChatGPT4o [18]. This is model of generating photorealistic portraits [19, 20] which perpetuate representational stereotypes [21].

Gender, ethnicity and age biases were found to be present in text-to-image creations and have been identified in DALL-E2 [5, 6, 22, 23], DALL-E3 [24], as well as the ChatGPT4 / DALL-E2 [25], ChatGPT4 / DALL-E3 [3, 26, 27] and ChatGPT4o / DALL-E3 combinations [4, 9]. More broadly, these biases are not limited to OpenAi products but can also be detected in other text-to-image applications such as Adobe Firefly, Stable Diffusion, and Midjourney [5, 7, 8, 28, 29].

The genesis of the observed biases rests in the nature and selection of the training data. These are exacerbated as cascading effects of decisions made during project development and subsequent red teaming phases [30, 31]. In a previous paper the author examined to what extent the observable biases of ethnicity, gender and age in the ChatGPT4o / DALL-E3 combination are due to autogenous prompt formulation in ChatGPT4o, or whether they were generated in the DALL-E3 diffusion model [32]. The paper showed that DALL-E3 introduces biases in all instances where ChatGPT-4o provided non-specific prompts.

Building on and extending that work, the aim of this brief study is to examine the prompt fidelity (prompt adherence, prompt compliance) of ChatGPT4o / DALL-E3 text-to-image visualisations. A dataset of autogenously generated prompts for the 200 visualisations of women working in cultural and creative industries allows us to examine whether specifications expressed in the ChatGPT4o autogenously prompts were actually implemented and visualised by the DALL-E engine [33].

## 2. Methodology

*2.1. Raw data*

2.1.1. Women working in cultural and creative industries

The data were drawn from the autogenously generated prompts of 200 visualisations of women working in cultural and creative industries contained in a public-domain dataset [33]. These prompts had been generated by ChatGPT4o in response to the following unconstrained prompt: "*Think about* [professional location] *and the* [professionals] *working in these. Provide me with a visualisation that shows a typical female* [professional] *against the background of the interior of the* [professional location]". Assessed were the following roles of women working in cultural and creative industries: actors in a theatre or playhouse; architects in an architectural design studio; artists in an artist's studio; authors in an author's creative space; curators in an art museum, fashion museum, natural history museum, science museum, and social history museum; librarians in a public library, school library, and university library; movie directors in a movie production set; musicians in an opera house or concert hall; and photographers in a photographic studio. Ten iterations were created for each of the roles [33]. Each resulting image was saved to disk, the Ai-generated prompt retrieved from the image panel (Figure 1) and, together with the image, copied into the data file. After saving, the chat was deleted to ensure a clean and unbiased generation of a new image without legacy information available to ChatGPT4o.

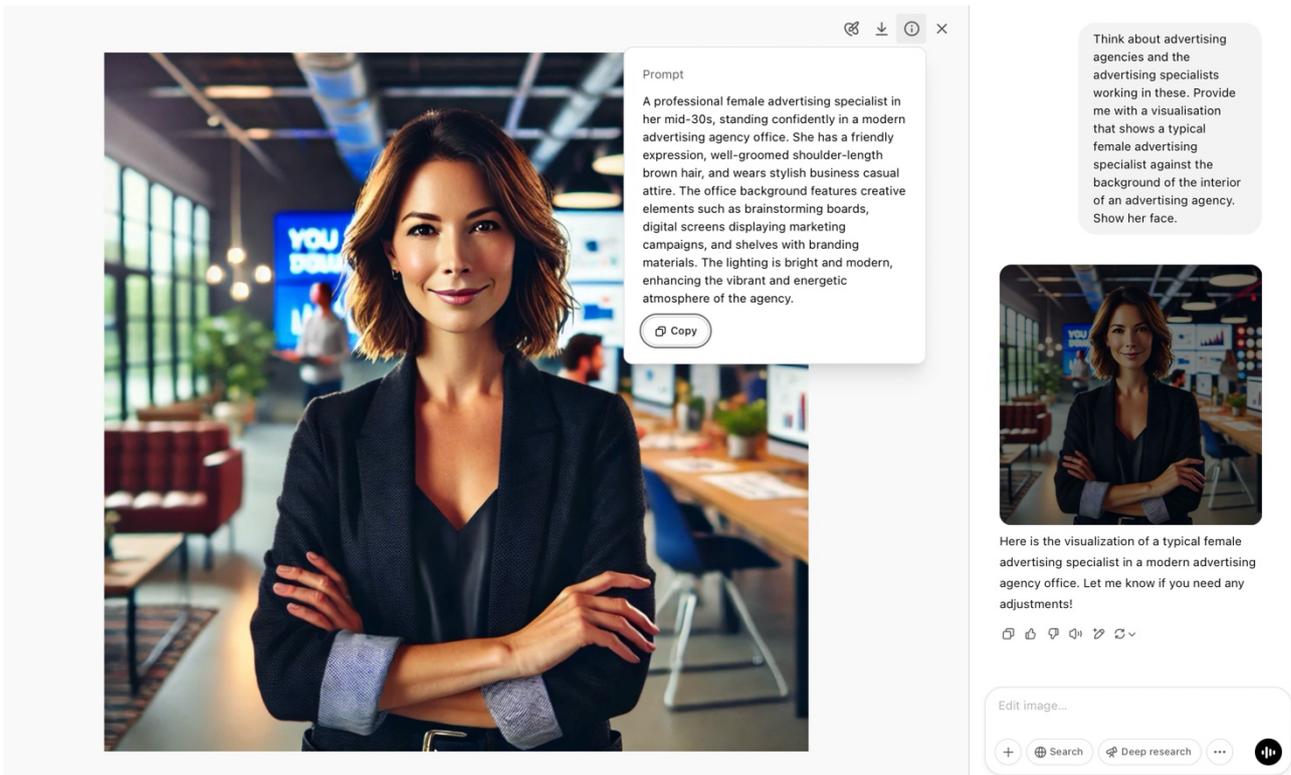

*Figure 1. Screen shot showing a ChatGPT4.5 image panel and DALL-E 'reading' of the prompt (pop-up window) together with text of the prompt as issued (right panel).*

2.1.2. Curators

The data were drawn from the autogenously generated prompts of 230 visualisations of curators contained in a public-domain dataset [34]. These prompts had been generated by ChatGPT4o in response to the following unconstrained prompt: "*Think about [insert museum type] museums and the curators working in these. Provide me with a visualisation that shows a typical curator against the background of the interior of the museum.*" The visualisations included fifty images each of curators depicted in art, fashion, natural history and social history museums and ten images each of curators depicted in maritime, science and technology museums. The process of saving and repeating the image generation was the same as that described above.

*2.2. Scoring*

Scored were visual cues to attire, hair style (bob, open long hair, bun, ponytail, etc) as well as the presence/absence of glasses (spectacles), a name tag, as well as a clipboard (or tablet). Age cohort scoring occurred via classification by ChatGPT4o using the following prompt: "Evaluate this image of a female professional woman. Examine the face and classify it for the age of the depicted woman. Classify the face in one of the following age classes '10 to 20 years of age', '20 to 30 years, '30 to 40 years, '40 to 50 years, '50 to 60 years, '60 to 70 years, and '70 to 80 years."

*2.3. Limitations*

As the two data sets were generated before the new ChatGPT native image generation model was rolled out in late March 2025, the observations made in this study may not reflect the current state of generative AI text-to-image visualisations. They do, however, serve as baseline data for comparison with the most recent models.

# 3. Results

*3.1. Women working in cultural and creative industries*

The instances where the prompts were silent of the specific attributes of age, hair, attire, glasses, name tags and clipboards (49.5% to 95.5%) were set aside. Where the prompts contained specific attributes, it was possible to ascertain whether the DALL-E3 rendered these attributes correctly. In the case of attire, DALL-E3 rendered some prompts correctly but added additional elements that were not specified in the prompt. These were scored as 'correct.' In the case of the attribute 'hair' some prompt specifications were open to interpretation (e.g. 'curly,' 'styled,' 'neatly styled'). To be conservative in the analysis, these were also scored as 'correct.'

The proportion of incorrectly rendered attributes ranged from 3.6% in the case of glasses (n=83) to 50% in the case of age (n=16) (Table 1). The combined proportion of incorrectly rendered attributes of the six sample pairs is 12.2% (n= 254). With the exception of the sample pair age and hair ($p=0.0362$, df=df 1, $\chi^2= 4.388$), there are no significant differences [35]. While the proportion of incorrectly rendered attributes appears to be higher among small samples (e.g. hair, name tag, age), there is no significant correlation ($r = -0.6243$) between the sample size per attribute and the proportion of incorrectly rendered attributes among the six sample pairs ($p = 0.1852$) [36].

*Table 1. Frequency of specifications by attribute of women working in cultural and creative industries (for detailed data see Table 3–Table 8).*

|  | Attribute | | | | | |
| --- | --- | --- | --- | --- | --- | --- |
|  | Age | Hair | Attire | Glasses | Name Tag | Clipboard |
| **Unspecified** | 184 | 166 | 99 | 117 | 191 | 189 |
| **Specified** | | | | | | |
| Correct | 8 | 21 | 56 | 80 | 6 | 10 |
| Correct (additional) *) | | | 36 | | | |
| Open to interpretation | | 6 | | | | |
| Incorrect | 8 | 7 | 9 | 3 | 3 | 1 |
| Total specified | 16 | 34 | 101 | 83 | 9 | 11 |
| % incorrect | 50.0 | 20.6 | 8.9 | 3.6 | 33.3 | 9.1 |

*) Correct with additional elements that were not specified in the prompt.

*3.2. Curators*

As above, the instances where the prompts were silent of the specific attributes of age, attire, glasses, name tags and clipboards (33.2% to 76.1%) were set aside. The proportion of incorrectly rendered attributes ranged from 4.5% in the case of glasses (n=89) to 38.3% in the case of attire (n=107)(Table 2). The combined proportion of incorrectly rendered attributes of the five sample pairs is 17.5% (n=456).

The incorrectly rendered attributes fall into two groups age and attire with proportions over 30% each and glasses, name tags and clipboards with proportions below 10% each. Very significant differences can be observed between the sample pairs clipboard and age ($p < 0.0001$, df=1, $\chi^2= 23.485$). Similar significant differences ($p < 0.0001$) exist with all other pairing of the two groups. While the proportion of incorrectly rendered attributes appears to be higher among small samples (e.g. hair, name tag, age), there is no significant correlation ($r = -0.1577$) between the sample size per attribute and the proportion of incorrectly rendered attributes among the five sample pairs ($p = 0.8001$) [36].

*Table 2. Frequency of specifications by attribute of curators*
*(for detailed data see Table 9-Table 13)*

|  | Attribute | | | | |
| --- | --- | --- | --- | --- | --- |
|  | Age | Attire | Glasses | Name Tag | Clipboard |
| **Unspecified** | 172 | 123 | 141 | 175 | 73 |
| **Specified** | | | | | |
| Correct | 30 | 66 | 85 | 51 | 136 |
| unclear | | | | 1 | |
| Incorrect | 20 | 41 | 4 | 4 | 11 |
| Total specified | 58 | 107 | 89 | 55 | 147 |
| % incorrect | 34.5 | 38.3 | 4.5 | 7.3 | 7.5 |

## 4. Discussion and Conclusions

While there is a considerable body of literature on prompt engineering in order to maximise user satisfaction with rendered images [37-39], only a limited number of papers have examined prompt fidelity (prompt adherence), and then only in very general terms [40-42]. These studies focussed on whether the overall scene composition complied with expectations, rather than whether individual, specific elements were actually included or excluded. Previous work by the author had shown that DALL-E3 may include unwanted image elements that were not included in the prompt text and that could not be excluded despite explicit prompt text specifications [30]. As discussed elsewhere, DALL-E3 has a considerable latitude in interpreting prompts that do not directly specify design elements [32].

A previous study of the ChatGPT4o / DALL-E3 text-to-image combination, which examined biases inherent in the images of 640 librarians and curators, found that 6.3% of images with a prompt-specified gender (n=79) deviated from the specification [32]. The present study is the first study that formally examines the issue of prompt fidelity of ChatGPT4o / Dall-E3 text-to-image visualisations.

By examining 200 visualisation of women working in cultural and creative industries and 230 visualisations of museum curators, the study confirmed that not all elements in ChatGPT4o autogenously generated prompt texts are being rendered by DALL-E3. The present study has shown that DALL-E3 incorrectly rendered an aggregate of 15.6% of specific prompt attributes (n=710) contained in ChatGPT4o autogenously generated prompts. The aggregate proportion of incorrectly rendered attributes was the lowest among paraphernalia (name tags, clipboards) with 6.6% (n=394). With 15.0% (n=380), attributes related to personal appearance (attire, glasses) were significantly more frequently incorrectly rendered ($p = 0.0002$, df=1, $\chi^2 = 14.278$). The greatest proportion of incorrectly rendered attributes (32.4%, n = 108) were those related to the depiction of a person's (age, hair) (personal vs. appearance attributes, $p < 0.0001$, df=1, $\chi^2 = 16.610$).

The findings highlight that even when prompts are explicitly specified, generative models selectively comply with attribute instructions, which undermines their reliability for controlled representation tasks. This has broader implications for bias auditing and transparency, as deviations in rendering personal attributes distort demographic accuracy and can perpetuate subtle stereotype leakage in downstream visual applications.

Table 3. Women working in cultural and creative industries. Correlation of ages specified in the prompt and their visualisations

| | Age definitions contained in the prompt | | | | | | | |
|---|---|---|---|---|---|---|---|---|
| | young | Late 20s or early 30s | mid-30s | late 30s or early 40s | mid-40s | middle-aged | un-specified | Total |
| 10–20 | | | | | | | 2 | 2 |
| 20–30 | 1 | 1 | 3 | | | 3 | 168 | 176 |
| 30-40 | | | 3 | 1 | 2 | | 13 | 19 |
| 40-50 | | | | | | 2 | | 2 |
| 50-60 | | | | | | | 1 | 1 |
| Total | 1 | 1 | 6 | 1 | 2 | 5 | 184 | 200 |

Table 4. Women working in cultural and creative industries. Correlation of elements of attire specified in the prompt and their visualisations

| observed | specified | | | | | | | | | | | | | |
|---|---|---|---|---|---|---|---|---|---|---|---|---|---|---|
| | blazer | blazer & blouse | blazer & jeans | blazer & t-shirt | blazer & skirt | cardigan | cardigan & blouse | dress | jacket & blouse | jean & sweater | jeans & blouse | jeans & t-shirt | unspecified | Grand Total |
| blazer & buttoned-up blouse | 1 | | | | | | | | | | | | 1 | 2 |
| blazer & open blouse | 16 | 22 | 1 | | 4 | | | | | | 1 | | 15 | 59 |
| blazer & t-shirt | | | | 3 | | | | | | | | | 2 | 5 |
| blouse | | | | | | | | | | | | | 6 | 6 |
| cardigan & buttoned-up blouse | | | | | 1 | 3 | 2 | | | | | | | 6 |
| cardigan & open blouse | | | | | | 5 | 9 | 16 | | | | | | 30 |
| cardigan & t-shirt | | | | | | 3 | 1 | | | | | | 1 | 5 |
| dress | | | | | | | | 1 | | | | | 20 | 21 |
| jacket & t-shirt | | | | | | | | | | | | | 1 | 1 |
| jacket, vest & button-up blouse | 1 | | | | | | | | | | | | | 1 |
| jacket, vest & open blouse | | | | | 1 | | | | | | | | | 1 |
| jacket, vest & t-shirt | | | | | | | | | | | | | 1 | 1 |
| jacket, vest & blouse | 1 | | | | | | | | | | | | | 1 |
| jumpsuit & blouse | | | | | | | | | | | | | 1 | 1 |
| labcoat & open blouse | | | | | | | | | | | | | 1 | 1 |
| open blouse | 1 | | | | | | | | | | 4 | | 20 | 25 |
| open blouse & apron | | | | | | | | | | | | | 5 | 5 |
| overall & open blouse | | | | | | | | | | | | | 1 | 1 |
| overall & open shirt | | | | | | | | | | | | | 2 | 2 |
| shirt & blouse | | | | | | | | | | | 1 | | 7 | 8 |
| shirt & t-shirt | | | | | | | | | 1 | | | 1 | 1 | 3 |
| sweater | | | | | | | | | | 1 | | | 8 | 9 |
| sweater & open blouse | | | | | | | | | | | | | 1 | 1 |
| t-shirt | | | | | | | | | | | | | 1 | 1 |
| t-shirt & apron | | | | | | | | | | | | | 3 | 3 |
| vest & open blouse | | | | | | | | | | | | | 1 | 1 |

| | | | | | | | | | | | | | |
|---|---|---|---|---|---|---|---|---|---|---|---|---|---|
| Total | | 20 | 22 | 1 | 3 | 11 | 15 | 19 | 1 | 1 | 1 | 6 | 1 | 99 | 200 |

Table 5. Women working in cultural and creative industries. Correlation of hair styles specified in the prompt and their visualisations

| | specified | | | | | | | | | | | |
|---|---|---|---|---|---|---|---|---|---|---|---|---|
| observed | classic updo | curly | messy bun | neat bun | neatly styled | neatly tied back | neatly tied up | short | shoulder-length | styled | unspecified | Total |
| bun | 1 | | | 6 | 1 | 3 | 1 | | | 1 | 53 | 66 |
| open long | | 1 | | | | 2 | | | 5 | | 55 | 63 |
| open shoulder | | | | | 2 | | | | 4 | | 39 | 45 |
| ponytail | | | | | | | | | | | 1 | 1 |
| short | | | | | 1 | | | 4 | | | 10 | 15 |
| tied back | | | | | | | | | | | 6 | 6 |
| tied up | | | 1 | | | | | | | 1 | 2 | 4 |
| Total | 1 | 1 | 1 | 7 | 3 | 5 | 1 | 4 | 9 | 2 | 166 | 200 |

Table 6. Women working in cultural and creative industries. Correlation of glasses specified in the prompt and their visualisations

| | specified | | |
|---|---|---|---|
| observed | yes | no | Total |
| yes | 80 | 12 | 92 |
| no | 3 | 105 | 108 |
| Total | 83 | 117 | 200 |

Table 7. Women working in cultural and creative industries. Correlation of name tags specified in the prompt and their visualisations

| | specified | | |
|---|---|---|---|
| observed | yes | no | Total |
| yes | 6 | 8 | 14 |
| no | 3 | 183 | 186 |
| Total | 9 | 191 | 200 |

*Table 8. Women working in cultural and creative industries. Correlation of clipboards or tablets specified in the prompt and their visualisations*

|  | specified |  |  |
| --- | --- | --- | --- |
| observed | yes | no | Total |
| yes | 10 | 20 | 30 |
| no | 1 | 169 | 166 |
| Total | 11 | 189 | 200 |

*Table 9. Curators. Correlation of ages specified in the prompt and their visualisations*

| observed | middle-aged | unspecified | Total |
| --- | --- | --- | --- |
| 10–20 | 3 | 98 | 94 |
| 20–30 | 16 | 72 | 88 |
| 30-40 | 30 |  | 30 |
| 40-50 | 9 | 1 | 10 |
| 50-60 |  | 0 | 0 |
| 70-90 |  | 1 | 1 |
| Total | 58 | 172 | 230 |

*Table 10. Curators. Correlation of elements of attire specified in the prompt and their visualisations*

| observed | blazer | cardigan or blazer | lab coat | lab coat or blazer | unspecified | Total |
| --- | --- | --- | --- | --- | --- | --- |
| ensemble, open shirt | 3 |  |  |  | 2 | 5 |
| jacket, open shirt | 38 | 2 |  | 2 | 39 | 81 |
| jacket, vest/sweater, open shirt | 12 | 1 |  |  | 17 | 30 |
| laboratory coat, open shirt |  |  |  | 1 | 1 | 2 |
| laboratory coat, shirt & tie | 1 |  | 1 | 2 | 5 | 9 |
| open shirt |  |  |  |  | 3 | 3 |
| shirt & tie |  |  |  |  | 6 | 6 |
| suit, shirt & tie | 7 |  |  | 1 | 16 | 24 |
| suit, vest & shirt | 3 |  |  |  | 1 | 4 |
| suit, vest, shirt & tie | 22 |  |  |  | 22 | 44 |
| vest, shirt & tie | 11 |  |  |  | 11 | 22 |
| Total | 97 | 3 | 1 | 6 | 123 | 230 |

*Table 11. Curators. Correlation of glasses specified in the prompt and their visualisations*

|  | specified |  |  |
| --- | --- | --- | --- |
| observed | yes | no | Total |
| yes | 85 | 24 | 108 |
| unclear |  | 2 | 2 |
| no | 4 | 115 | 113 |
| Total | 89 | 141 | 230 |

*Table 12. Curators. Correlation of name tags specified in the prompt and their visualisations*

|  | specified | | |
|---|---|---|---|
| observed | yes | no | Total |
| yes | 50 | 14 | 64 |
| unclear | 1 | 5 | 6 |
| no | 4 | 156 | 160 |
| Total | 55 | 175 | 230 |

*Table 13. Curators. Correlation of clipboards or tablets specified in the prompt and their visualisations*

| observed | specified | | | | | | | | |
|---|---|---|---|---|---|---|---|---|---|
|  | clipboard | clipboard or notebook | clipboard or tablet | notebook | notebook or magnifying glass | notebook or tablet | tablet | unspecified | Total |
| book |  | 1 |  | 1 |  |  |  |  | 2 |
| clipboard | 114 | 3 | 11 | 7 |  | 1 |  | 32 | 168 |
| clipboard & camera |  |  |  |  |  |  |  | 1 | 1 |
| none | 1 |  |  |  |  |  |  | 30 | 31 |
| notebook | 2 |  |  | 3 | 1 |  |  |  | 6 |
| object |  |  |  |  |  |  |  | 4 | 4 |
| tablet |  |  | 7 |  |  | 1 | 4 | 6 | 18 |
| Total | 117 | 4 | 18 | 11 | 1 | 2 | 4 | 73 | 230 |

**Funding**: This research received no external funding.

**Institutional** Review Board Statement: Not applicable

**Conflicts of Interest**: The author declares no conflict of interest.